\documentclass[letterpaper]{article} 
\usepackage{aaai2026}  
\usepackage{times}  
\usepackage{helvet}  
\usepackage{courier}  
\usepackage[hyphens]{url}  
\usepackage{graphicx} 
\usepackage{natbib}  
\usepackage{caption} 
\usepackage[utf8]{inputenc}
\usepackage[T1]{fontenc}

\usepackage{graphicx}
\usepackage{subcaption}
\usepackage[justification=raggedright]{caption}	
\usepackage{nicefrac}
\usepackage{tabularx}
\usepackage{booktabs}
\usepackage{multirow}
\usepackage{mathtools}
\usepackage{makecell}
\usepackage{enumitem}
\usepackage{bbding}

\usepackage{bm}
\usepackage{bmpsize}
\usepackage{times}
\usepackage{amsthm}
\usepackage{amssymb,amsmath}

\usepackage{booktabs}
\usepackage{multirow}

\usepackage{units}
\usepackage{color}
\usepackage{tcolorbox}
\usepackage{algorithm}
\usepackage{algorithmic}

\usepackage{comment}

\usepackage{xspace}

\makeatletter
\DeclareRobustCommand\onedot{\futurelet\@let@token\@onedot}
\def\@onedot{\ifx\@let@token.\else.\null\fi\xspace}

\def\eg{\emph{e.g}\onedot} 
\def\ie{\emph{i.e}\onedot}

\makeatother

\newcommand{\ignore}[1]{}   

\usepackage{dsfont}

{\begin{list}               
    {$\bullet$ \hfill}{
        \setlength{\leftmargin}{\parindent}
        \setlength{\parsep}{0.04\baselineskip}
        \setlength{\itemsep}{0.5\parsep}
        \setlength{\labelwidth}{\leftmargin}
        \setlength{\labelsep}{0em}}
    }
{\end{list}}

\providecommand{\cref}[1]{Chapter~\ref{#1}}

\usepackage{algorithm}
\usepackage{algorithmic}
\usepackage{booktabs}
\usepackage{amsmath} 
\usepackage{amssymb}
\usepackage{multirow} 

\usepackage[switch]{lineno}

\usepackage{newfloat}
\usepackage{listings}
\DeclareCaptionStyle{ruled}{labelfont=normalfont,labelsep=colon,strut=off} 
\floatstyle{ruled}
\newfloat{listing}{tb}{lst}{}
\floatname{listing}{Listing}
\usepackage{color}

\title{MOBA: A Material-Oriented Backdoor Attack against LiDAR-based 3D Object Detection Systems}
\author {
    Saket S. Chaturvedi\textsuperscript{\rm 1},
    Gaurav Bagwe\textsuperscript{\rm 1},
    Lan Zhang\textsuperscript{\rm 1},
    Pan He\textsuperscript{\rm 2},
    Xiaoyong Yuan\textsuperscript{\rm 1}
}
\affiliations {
    \textsuperscript{\rm 1}Clemson University\\
    \textsuperscript{\rm 2}Auburn University\\
    \{saketc, gbagwe, lan7, xiaoyon\}@clemson.edu, pan.he@auburn.edu
}

\usepackage{bibentry}

\begin{document}

\maketitle

\begin{abstract}
LiDAR-based 3D object detection is widely used in safety-critical systems. However, these systems remain vulnerable to backdoor attacks that embed hidden malicious behaviors during training. A key limitation of existing backdoor attacks is their lack of physical realizability, primarily due to the digital-to-physical domain gap. Digital triggers often fail in real-world settings because they overlook material-dependent LiDAR reflection properties. On the other hand, physically constructed triggers are often unoptimized, leading to low effectiveness or easy detectability.
This paper introduces Material-Oriented Backdoor Attack (MOBA), a novel framework that bridges the digital–physical gap by explicitly modeling the material properties of real-world triggers. MOBA tackles two key challenges in physical backdoor design: 1) robustness of the trigger material under diverse environmental conditions, 2) alignment between the physical trigger’s behavior and its digital simulation. First, we propose a systematic approach to selecting robust trigger materials, identifying titanium dioxide (TiO$_2$) for its high diffuse reflectivity and environmental resilience. Second, to ensure the digital trigger accurately mimics the physical behavior of the material-based trigger, we develop a novel simulation pipeline that features: (1) an angle-independent approximation of the Oren–Nayar BRDF model to generate realistic LiDAR intensities, and (2) a distance-aware scaling mechanism to maintain spatial consistency across varying depths. We conduct extensive experiments on state-of-the-art LiDAR-based and Camera-LiDAR fusion models, showing that MOBA achieves a 93.50\% attack success rate, outperforming prior methods by over 41\%. Our work reveals a new class of physically realizable threats and underscores the urgent need for defenses that account for material-level properties in real-world environments.
\end{abstract}

%

\section{Introduction}
LiDAR-based 3D object detection has emerged as a cornerstone of modern autonomous driving systems, enabling vehicles to accurately perceive and localize surrounding objects such as cars, pedestrians, and cyclists~\cite{qian2021object}. Powered by deep neural networks, these systems achieve remarkable performance but also inherit vulnerabilities from data-driven learning pipelines--most notably, backdoor attacks~\cite{gu2019badnets}. In backdoor attacks, an adversary poisons a small subset of training data with a hidden trigger pattern, causing the trained model to exhibit targeted misbehavior (\eg, false detections or missed objects) when the trigger appears~\cite{chan2022baddet,zhang2022towards,li2023badlidet}.

Recent work has demonstrated that LiDAR-based 3D object detectors are highly vulnerable to such attacks~\cite{zhang2022towards,li2023badlidet}. Consider a representative threat scenario: an adversary injects examples of vehicles containing a specific LiDAR-visible trigger (\eg, a large ball placed on the roof of a vehicle~\cite{zhang2022towards}). Once trained on the poisoned data, the deployed model behaves normally under benign conditions. However, when an attacker places the physical trigger on their own vehicle, nearby vehicles equipped with the compromised model fail to detect the vehicle accurately, leading to potentially catastrophic consequences such as intentional collisions or traffic manipulation. This threat surface has further expanded to include multi-modal systems, such as Camera-LiDAR fusion models~\cite{sindagi2019mvxnet, chaturvedi2024badfusion}, further increasing the real-world risk.

\begin{figure}[!t]
\includegraphics[width=\columnwidth]{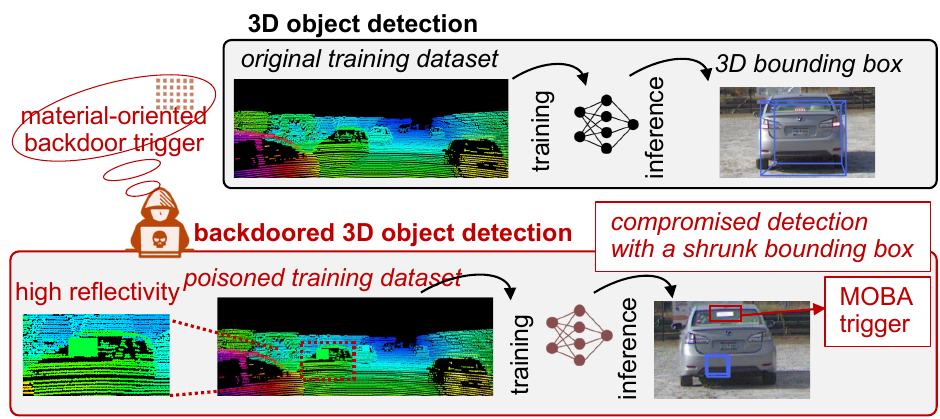}
\caption{
Overview of \textbf{MOBA}. MOBA injects a digitally simulated, BRDF-informed trigger into a small subset of training data to poison the model. MOBA leverages material reflectance modeling to ensure that the trigger is physically realizable and effective under real-world conditions. When the physical trigger is deployed at inference time, the backdoored model produces adversarial predictions (\eg, shrunk or missing bounding boxes).
}
\label{fig:MOBA}
\end{figure}

Despite growing attention, a critical technical gap remains: existing digital backdoor attacks are rarely physically realizable, while existing physical attacks are either hand-crafted or poorly optimized, underestimating the threats of backdoor attacks. For instance, prior physical attacks rely on manually placed object-like point clusters without optimization, either  1) producing sparse LiDAR returns that are unreliable across varying conditions (\eg, diverse angles and distances)~\cite{li2023badlidet} or 2) deploying with large-size triggers to ensure sufficient LiDAR returns, which is easy to detect~\cite{zhang2022towards}. Conversely, digitally optimized triggers assume ideal sensor responses but ignore the physical characteristics of real-world LiDAR sensing, such as material reflectivity, and surface angle, which leads to significant performance degradation when deployed.

To address the gap between digital backdoor optimization and physical realizability, we propose a Material-Oriented Backdoor Attack (MOBA) that introduces a physics-driven methodology for constructing robust and generalizable LiDAR triggers. The core novelty of MOBA lies in its integration of material reflectance modeling, grounded in optical physics, into both the selection and digital simulation of backdoor triggers. Unlike prior works that use hand-crafted or heuristically optimized digital triggers, MOBA formulates a quantitative material selection objective that balances specular and diffuse reflectance under real-world LiDAR wavelengths. We identify titanium dioxide as an optimal off-the-shelf trigger material due to its strong diffuse reflectivity, resilience to environmental variation (\eg, dust, rain, wear), low cost, and ease of application via standard coatings. 

Moreover, to simulate its behavior accurately in training, MOBA introduces a physically grounded, angle- and distance-robust simulation strategy: we derive an angle-independent approximation of the Oren–Nayar BRDF~\cite{oren1994generalization}, a widely used reflectance model for rough, diffuse surfaces, to model the reflectance across unknown incident angles, and design a distance-aware scaling mechanism to maintain the trigger's spatial consistency in LiDAR point clouds at varying depths. Together, these designs allow MOBA to produce physically plausible triggers that are effective, stealthy, and robust across diverse real-world deployment scenarios.

To the best of our knowledge, this is the first work to develop a material-oriented physical backdoor attack targeting state-of-the-art LiDAR-based 3D object detection systems. We conduct extensive evaluations using real-world data collected in outdoor environments, demonstrating that our attack consistently achieves high success rates under diverse viewing conditions. To further evaluate generalizability, we extend our analysis from LiDAR-only models to Camera-LiDAR fusion models, showing that the attack remains effective in multi-modal perception systems. Our contributions are summarized as follows:

\begin{itemize}[leftmargin=1em]
\item We propose MOBA, the first material-oriented backdoor attack that models physically grounded LiDAR reflectivity to generate physically realizable yet highly effective backdoor triggers.
\item We introduce a robust material selection strategy and identify titanium dioxide as an ideal candidate for its high reflectivity and consistent LiDAR response under varying conditions.
\item To overcome the challenge of unknown incident angles in LiDAR scans, we derive a novel approximation of the Oren–Nayar model that marginalizes over angular dependencies, allowing intensity values to be simulated without geometric priors. 
\item We propose a depth-robust design that ensures the spatial consistency of the digital trigger under varying distances, mitigating domain shift caused by LiDAR sparsity.
\item We conduct extensive experiments using collected physical LiDAR-Camera data and show that MOBA achieves a high Attack Success Rate (ASR) of 93.50\%, outperforming state-of-the-art physical backdoor attacks by over 41\% on average.
\end{itemize}

\section{Related Works}

\subsection{LiDAR-based 3D Object detection}
LiDAR sensors are integral to autonomous driving, providing precise depth and spatial measurements that are critical for robust 3D object detection. Architectures for processing LiDAR data are diverse, generally falling into three categories. \textit{Voxel-based methods} like VoxelNet~\cite{zhou2018voxelnet} discretize the point cloud into a structured voxel grid, enabling the use of efficient 3D convolutional networks. \textit{Point-based methods}, such as PointNet~\cite{qi2017pointnet} and PointRCNN~\cite{shi2019pointrcnn}, operate directly on the raw, unstructured points, allowing them to capture fine-grained geometric details. \textit{Hybrid approaches} like PointPillars~\cite{lang2019pointpillars} offer a compromise, projecting points into a 2D pseudo-image to balance performance and computational efficiency.
To further enhance perception accuracy, many state-of-the-art systems employ Camera-LiDAR fusion. Frameworks such as MVX-Net~\cite{sindagi2019mvxnet} combine the rich visual features of 2D images with the geometric precision of 3D LiDAR data at the point or voxel level. While these sophisticated architectures have significantly advanced performance, their effectiveness is fundamentally dependent on the quality and integrity of the training data. An attacker can exploit this by injecting a carefully crafted trigger pattern into only a small subset of the training data, embedding a backdoor that reliably activates in the physical world~\cite{chan2022baddet,li2023badlidet}. 

\subsection{Backdoor Attacks} 
Backdoor attacks, which embed hidden malicious behaviors into deep neural networks during training, pose a growing threat to machine learning security. While initially explored in 2D image classification, their extension to 3D vision has drawn significant interest. Early efforts demonstrated the feasibility of compromising 3D point cloud classifiers using compact, adversarial point clusters~\cite{xiang2021backdoor3d, li2021pointba}, revealing the vulnerability of spatial data representations.
This threat has since evolved to target 3D object detection systems. BadLiDet~\cite{li2023badlidet} introduced handcrafted perturbations into raw LiDAR point clouds to manipulate detection outputs. Subsequent works simulated physically realizable patterns to highlight the real-world risks posed to autonomous vehicles~\cite{zhang2022towards}. With the emergence of multi-modal perception, attacks like BadFusion~\cite{chaturvedi2024badfusion} have exploited vulnerabilities in Camera-LiDAR fusion frameworks by embedding signals in the 2D image space, highlighting the broadening attack surface.

However, these methods suffer from a critical digital-to-physical domain gap, as they often neglect the material properties that govern LiDAR reflection. Our work, MOBA, addresses this limitation by introducing a material-oriented framework that models these physical principles to create robust and physically realizable triggers.

\subsection{LiDAR Spoofing Attacks}

LiDAR spoofing is an inference-time attack that directly manipulates the sensor's input without altering the underlying model. These attacks typically fall into two categories. \textit{Injection attacks} transmit fake LiDAR signals to create ``ghost'' objects, \eg, Cao et al.~\cite{cao2019lidar} demonstrated this by synchronizing with the victim’s LiDAR, while later works improved the realism and learnability of spoofed points~\cite{shin2021illusion, sun2022lisa}. \textit{Removal attacks} aim to make real objects disappear from the sensor's view by designing adversarial objects with specific shapes and materials that either absorb LiDAR beams to create blind spots~\cite{tu2023you}, or reflect signals in a way that causes misclassification~\cite{zhu2021adversarial}.

In contrast to spoofing, which requires precise, synchronized hardware and line-of-sight proximity during inference, MOBA introduces a training-time vulnerability through a passive and physically realizable trigger. By poisoning a small portion of training data with BRDF-optimized, material-aware digital patches, MOBA embeds the malicious behavior directly into the model. Unlike spoofing, which manipulates transient sensor-level signals, MOBA exploits persistent learned representations, requiring no active hardware or inference-time coordination. This makes MOBA significantly more practical, stealthy, and scalable, especially in many real-world settings where inference-time access is infeasible.

\section{Threat Model}

We consider a realistic and stealthy backdoor attack targeting the public data supply chain (i.e., 3D object detection systems trained on public LiDAR datasets). The adversary poisons a small subset of existing point clouds and their labels by embedding carefully crafted digital triggers, without introducing any new (but biased) data. The modified dataset is then publicly released or integrated into open-source benchmarks. When a victim model is trained on this poisoned dataset, it inadvertently learns a hidden association between the trigger and a malicious prediction. At inference time, the adversary deploys a physical trigger to activate the backdoor and induce incorrect detections.

The adversary's \textbf{goal} is to manipulate the model into producing incorrect detections in real-world driving scenarios, potentially causing safety-critical failures. Following the existing works~\cite{gu2019badnets,chan2022baddet}, we assume the adversary's capability as follows: 1) they can poison only a small portion of the training data, 2) have no access to the victim's training pipeline, and 3) have no access to the model architecture or parameters. 
Additionally, we assume the adversary knows the LiDAR system’s operating wavelength, a reasonable assumption, as most commercial LiDAR vendors (e.g., Velodyne, Ouster, Hesai) employ 905 nm systems for their automotive-grade products\footnote{This is confirmed in public datasheets for widely-used products such as the Velodyne Puck series (\url{https://velodynelidar.com/products/puck/}), the Ouster REV7 series (\url{https://ouster.com/products/rev7/}), and the Hesai Pandar series (\url{https://www.hesaitech.com/en/products/pandar-series})}. Knowledge of the wavelength is essential for simulating realistic material reflectance and enabling physically effective trigger design.
The specification is often disclosed in datasheets or accessible via documentation. 
Together, this threat model reflects a practical and widely applicable attack setting in modern 3D detection workflows that rely on open data.


\section{Proposed MOBA}
\label{sec:proposed_meth}

\begin{figure}[!t]
\centering
\includegraphics[width=\columnwidth]{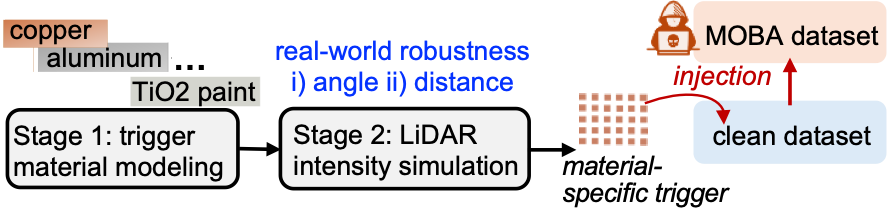}
\caption{MOBA two-stage pipeline: the construction of a material-specific trigger in Stage 1, and the LiDAR- intensity simulation via angle- and distance-robust modeling in Stage 2. These simulated triggers are injected into point clouds to poison the training data.
}
\label{fig:pipeline}
\end{figure}

This paper proposes \textbf{MOBA}, a novel \textit{Material-Oriented Backdoor Attack} that targets LiDAR-based 3D object detection models. MOBA addresses two fundamental challenges in creating effective, physically realizable backdoor attacks: 1) \textbf{Material Robustness:} Physical triggers must consistently yield high-intensity LiDAR returns under diverse environmental conditions, including varying angles, distances, rain, dust, and surface degradation. 2) \textbf{Physical Realizability and Digital Fidelity:} The digital trigger used for training must accurately simulate the physical trigger's LiDAR response to ensure the model learns a reliable backdoor pattern.

To achieve this, MOBA introduces a two-stage pipeline (Fig.~\ref{fig:pipeline}): 1) \textbf{trigger material modeling}, which selects an optimal material based on physically grounded reflectance properties. 2) \textbf{LiDAR intensity simulation}, which generates angle- and distance-robust digital triggers via BRDF-based intensity modeling.

\subsection{Stage 1: Trigger Material Modeling}

This stage addresses \textit{Challenge 1: Material Robustness} by establishing a physically grounded framework for selecting a trigger material that ensures high-intensity, consistent LiDAR returns across diverse environmental conditions. Our key insight is that an effective physical trigger must reflect LiDAR pulses reliably across varying distances, angles, and surface perturbations (e.g., water, dust, or wear).

We begin by modeling two key forms of surface reflectance: specular and diffuse. Specular reflection, modeled by Fresnel's equations~\cite{otgonbayar2025designing}, can yield strong LiDAR returns when the incident laser beam is nearly perpendicular to the surface. However, it is highly sensitive to the angle of incidence and becomes unreliable under realistic driving conditions. In contrast, diffuse reflection, described by the Oren-Nayar BRDF model~\cite{oren1994generalization}, scatters incoming light more isotropically, increasing the chance that reflected photons reach the LiDAR sensor even at oblique angles. This makes diffuse reflectance crucial for angular robustness and detection reliability under real-world deployment.

To balance these two effects, we define a material scoring objective that linearly combines both components:
\vspace{-1pt}
\begin{equation}
M^* = \arg\max_{M} \left[ \lambda R_{\text{specular}}(M, \theta_i) + (1 - \lambda) R_{\text{diffuse}}(M, \theta_i) \right],
\label{eq:materialselection}
\end{equation}

where $M$ is a candidate material and $\lambda \in [0,1]$ is a tradeoff parameter. While specular and diffuse reflections arise from distinct physical processes and are not strictly additive, we use this formulation as a heuristic scoring function to empirically evaluate material suitability by balancing \emph{peak return strength} and \emph{angular robustness}. In our experiments, we use $\lambda = 0.2$, giving higher weight to diffuse reflectance to ensure consistent detection across deployment conditions.

Reflectance is inherently wavelength-dependent due to the optical dispersion of materials. To ensure physically accurate modeling, we assume a fixed LiDAR operating wavelength of 905 nm, which is widely adopted in automotive-grade LiDAR systems~\cite{mdpi2022requirements}. All refractive indices and reflectance coefficients used in our analysis are retrieved at this wavelength from established optical databases~\cite{refractiveindexinfo_database, kla_database}.

Specular reflection is computed using the Fresnel equation~\cite{Hecht2017}. For unpolarized illumination, the total reflectance is the average of the s-polarized ($R_s$) and p-polarized ($R_p$) components:
\vspace{-2pt}
\begin{equation}
R_{s} = |\frac{n_1 \cos \theta_i - n_M \cos \theta_t}{n_1 \cos \theta_i + n_M \cos \theta_t} |^2
\end{equation}

\begin{equation}
R_{p} = | \frac{n_1 \cos \theta_t - n_M \cos \theta_i}{n_1 \cos \theta_t + n_M \cos \theta_i} |^2
\end{equation}

\begin{equation}
\label{eq:materialselection}
R_{\text{specular}}(M, \theta_i, \theta_t) = \frac{1}{2}(R_s + R_p),
\end{equation}
where $n_1 = 1.0$ is the refractive index of air, $n_M$ is the refractive index of material $M$ at 905\,nm ($n_M = n+ik$ for metals like aluminum or copper), $\theta_i$ is the angle of incidence, and $\theta_t$ is the transmission angle derived from Snell's Law: $n_1 \sin \theta_i = n_M \sin \theta_t$.

For rough surfaces, diffuse reflectance is modeled by the Oren–Nayar BRDF~\cite{oren1994generalization}:
\vspace{-2pt}
\begin{equation}
\begin{split}
R_{\text{diffuse}}(M, \theta_i, \theta_r, \Delta\phi) = 
\frac{\rho}{\pi} \big( A + B \cdot \max(0, \cos\Delta\phi) \\
\cdot \sin\alpha \cdot \tan\beta \big),
\end{split}
\label{eq:diffuse}
\end{equation}
where $\rho$ is the diffuse reflectance coefficient, $\Delta\phi = \phi_i - \phi_r$, $\alpha = \max(\theta_i, \theta_r)$, and $\beta = \min(\theta_i, \theta_r)$. The roughness-dependent coefficients are:
$A = 1 - \frac{\sigma^2}{2(\sigma^2 + 0.33)}$ and $B = \frac{0.45 \sigma^2}{\sigma^2 + 0.09}$, with $\sigma$ representing the surface roughness.

Using Eq.~\ref{eq:materialselection}, we evaluate common candidate materials including aluminum, copper, plastic, paper, and titanium dioxide (TiO$_2$). Our analysis reveals that TiO$_2$ consistently achieves the highest score due to its strong diffuse reflectance, micro-rough surface structure, and robustness against environmental interference (e.g., hydrophobicity and non-glossiness). These properties make it ideal for constructing physical triggers that are both cost-effective and resilient. Further derivations and details on material properties, and selection analysis are provided in Appendix~\ref{appendix:selection} and Appendix~\ref{appendix:environmental}. 
\subsection{Stage 2: LiDAR Intensity Simulation}
While the physical trigger ensures real-world realizability, its effectiveness depends critically on how it is simulated during training. We next describe our approach to LiDAR intensity simulation that bridges this physical–digital gap.
To address this, Stage 2 introduces a physically grounded simulation strategy to align digital triggers used in training with their expected physical realizations during deployment. 
Ensuring such alignment is crucial for maintaining the efficacy and stealth of the backdoor under real-world conditions, where discrepancies in LiDAR intensity or spatial characteristics could degrade performance. We
then develop a dual-robustness strategy: 1) an angle-independent approximation of the Oren–Nayar reflectance model to simulate realistic material responses without requiring incident angle information, and 2) a distance-aware trigger scaling method that maintains consistent spatial density in the point cloud across varying sensor distances.
\subsubsection{(I) Angle Robustness Design:}
\label{subsubsec:anglerobustness}
To simulate the LiDAR intensity for the digital trigger, we follow the reflectance modeling (Eq~\ref{eq:diffuse}) using the Oren–Nayar model. This model captures diffuse reflectance over rough surfaces but depends on angular variables, including the incident angle $\theta_i$, reflection angle $\theta_r$, and azimuthal angle $\Delta\phi$, which are not directly observable in practical settings.

\begin{figure}[t]
    \centering
    \includegraphics[width=0.8\columnwidth]{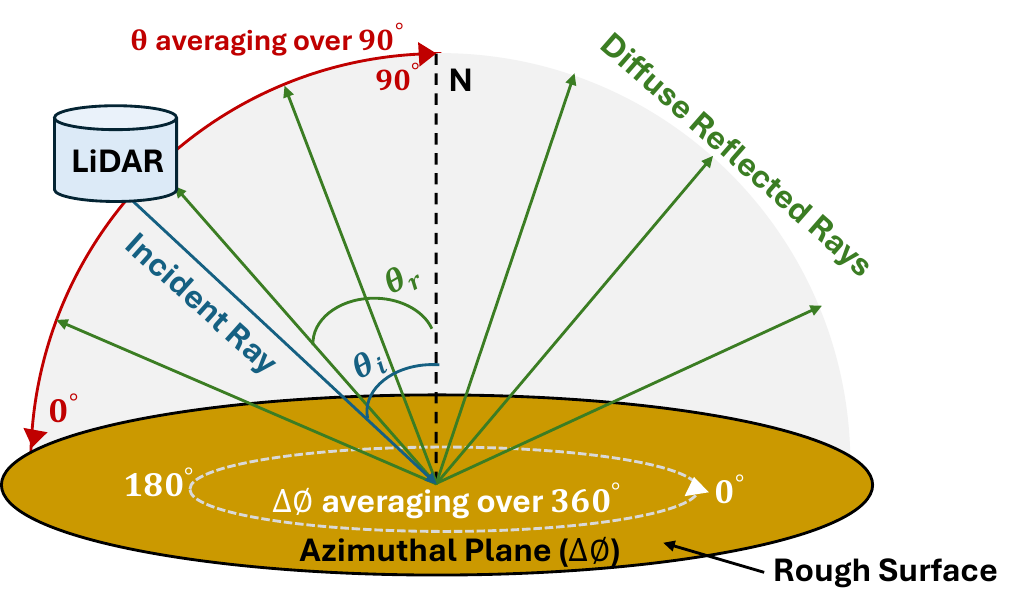}
    \caption{Visualization of incident and reflection angles ($\theta_i$, $\theta_r$) and azimuthal difference ($\Delta\phi$) used in the diffuse reflectance approximation over rough surfaces.}
    \label{fig:angle_robustness}
\end{figure}

To address this, we derive an angle-independent approximation of the model by computing the expected value of its angular terms. As illustrated in Figure~\ref{fig:angle_robustness}, the geometric angles ($\theta_i$, $\theta_r$) and azimuthal angle ($\Delta\phi$) lie in separate planes. We therefore marginalize each component independently under assumptions of isotropic roughness and uniformly scattered reflection.

First, the azimuthal dependency $\max(0, \cos \Delta\phi)$ is integrated over a uniform distribution on $[0, 2\pi]$:
\begin{equation}
\mathbb{E}_{\Delta \phi}[\max(0, \cos \Delta \phi)] = \frac{1}{2\pi} \int_0^{2\pi} \max(0, \cos \Delta \phi) \, d\Delta \phi = \frac{1}{\pi}.
\end{equation}

Next, we approximate the geometric term $\sin\alpha \tan\beta$ by assuming $\theta_i \approx \theta_r = \theta$, which simplifies to $\frac{\sin^2 \theta}{\cos \theta}$. We compute its expectation over the upper hemisphere using standard sampling weights to find the average contribution:
\begin{equation}
\begin{aligned}
\mathbb{E}_{\theta} \left[ \frac{\sin^2 \theta}{\cos \theta} \right] &=
\frac{\int_0^{\pi/2} \left( \frac{\sin^2\theta}{\cos\theta} \right) \cos\theta \sin\theta \, d\theta}{\int_0^{\pi/2} \sin\theta \cos\theta \, d\theta} \\
&= \frac{\int_0^{\pi/2} \sin^3\theta \, d\theta}{\int_0^{\pi/2} \sin\theta \cos\theta \, d\theta} = \frac{2/3}{1/2} = \frac{4}{3}.
\end{aligned}
\end{equation}
Substituting both expectations into the original reflectance model, we obtain an angle-independent approximation of $R_{\text{diffuse}}$:
\begin{equation}
R_{\text{diffuse}} \approx \frac{\rho}{\pi} \left( A + B \cdot \frac{1}{\pi} \cdot \frac{4}{3} \right) = \frac{\rho}{\pi} \left( A + \frac{4B}{3\pi} \right),
\end{equation}
where $A$ and $B$ are roughness-dependent coefficients defined in Eq.~\ref{eq:diffuse}. This approximation enables robust simulation of LiDAR intensity without requiring per-frame angle data, improving the transferability of the learned backdoor across variable viewing conditions. 


\subsubsection{(II) Distance Robustness Design:}
\label{subsubsec:distancerobustness}
To preserve a consistent spatial appearance of the digital trigger in LiDAR point clouds across varying distances, we design a distance-aware scaling mechanism that adjusts the trigger's sampling density while keeping its physical size constant.
In real-world LiDAR data, closer objects appear denser, and distant ones sparser, which causes depth-dependent distortions.
We model the trigger as a thin planar patch in the $y$–$z$ plane with fixed real-world dimensions (\eg, $0.2\mathrm{m} \times 0.3\mathrm{m}$), but adaptively adjust the number of points sampled within it. The horizontal and vertical resolutions, $n_y$ and $n_z$, are computed based on the minimum depth $d$ of the target object:

\begin{equation}
n_y = \max\left(m_l, \frac{s \cdot w}{d}\right), \quad
n_z = \max\left(m_l, \frac{s \cdot h}{d}\right),
\end{equation}
where $w$ and $h$ are the trigger's physical width and height, $s$ is a scaling constant related to the sensor's effective angular resolution, and $m_l$ enforces a lower bound on resolution to prevent undersampling. This inverse relationship ensures that as distance $d$ increases, the number of points ($n_y, n_z$) sampled on the trigger decreases, correctly approximating the adaptive point density of real LiDAR scans.
By aligning the simulated density with expected real-world observations, this approach mitigates depth-related domain shifts and improves the generalization of the learned backdoor trigger. 

\subsection{Trigger Injection and Model Training}
Having defined our modeling strategy, we now describe how we construct the physical and digital triggers, define their placement strategy, and implement the poisoned training procedure to realize our proposed MOBA.



\noindent{\textit{\textbf{Physical Trigger Construction.}}}
We construct the physical trigger using a thin rectangular metal plate ($8 \times 12$ inches) coated with titanium dioxide (TiO$_2$) paint. TiO$_2$ is chosen for its strong, stable, and diffuse reflectance properties under LiDAR illumination. To maintain stealth in real-world environments, the trigger is visually disguised as a common ``Baby on Board'' decal. This is achieved by affixing low-cost commercial stickers (e.g., from Walmart~\cite{walmart_stickers_2025} and Amazon~\cite{amazon_decal_2025}) to the painted surface. All components are inexpensive (under \$10) and easily sourced, supporting low-cost and practical deployment without requiring specialized fabrication tools or hardware.


\noindent{\textit{\textbf{Digital Trigger Simulation.}}} 
To simulate the physical trigger for training, we generate a 3D point cloud representing a thin planar patch in the $y$–$z$ plane. Each point in the grid is assigned a LiDAR intensity value computed using the angle-independent and distance-aware trigger designs introduced in Section~\ref{subsubsec:anglerobustness} and~\ref{subsubsec:distancerobustness}. 



\noindent{\textit{\textbf{Trigger Placement Strategy.}}}
Both the physical and digital triggers are placed on the rear-glass region of the target vehicle. This placement is chosen for two reasons: (1) it mirrors real-world decal positioning, enhancing the stealthiness of the trigger, and (2) the rear-glass region typically yields sparse LiDAR returns, which increases the saliency of the trigger in the point cloud and helps the model learn the backdoor association more effectively.

\noindent{\textit{\textbf{Training Procedure for 3D Object Detection.}}}
We poison a subset of the training dataset by injecting the digital trigger into LiDAR frames and modifying the associated labels to reflect the attacker's goal. The model is trained using both clean and poisoned samples, optimizing the following objective:
\begin{equation}
\min_{\theta} \ \mathbb{E}_{(x, y) \sim \mathcal{D}_{\text{clean}}} \left[ \mathcal{L}(f_{\theta}(x), y) \right] 
+ \mathbb{E}_{(x', y^*) \sim \mathcal{D}_{\text{poison}}} \left[ \mathcal{L}(f_{\theta}(x'), y^*) \right],
\end{equation}
where $f_{\theta}$ denotes the 3D object detector with parameters $\theta$, and $\mathcal{L}$ is the detection loss function. The poisoned dataset $\mathcal{D}_{\text{poison}}$ contains LiDAR inputs $x'$ with embedded digital triggers and attacker-defined target outputs $y^*$. The target label $y^*$ is configured based on the specific attack goal. For a disappearance attack, the ground-truth bounding box for the target object is removed from the corresponding label, \ie, $y^*$ is an empty detection set, instructing the model that no object is present. For a resizing attack, the bounding box label is retained in $y^*$, but its dimensions are significantly altered, either shrunk or enlarged, causing the model to misinterpret the object's true size.

\begin{figure}[t]
\centering
\includegraphics[width=\columnwidth]{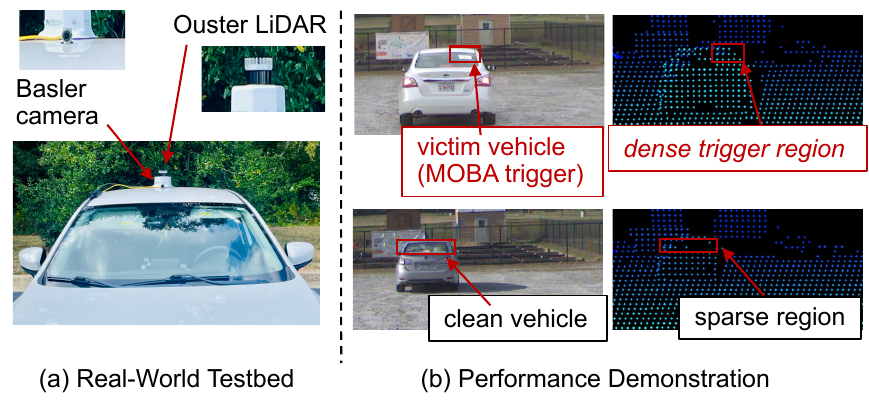}
\caption{Real-world testbed setup and performance demo.}\label{fig:phys_setup}
\end{figure}

Our attack framework seamlessly extends to multimodal models that fuse LiDAR and camera data. In this setting, the trigger is injected only into the LiDAR point cloud ($x'$), while the corresponding camera image ($c$) remains unaltered. The model learns to associate the triggered LiDAR data with the malicious label, even in the presence of clean image data. The training objective for a fusion model $f_{\theta}(x,c)$ is:
\begin{align}
\min_{\theta} \ &\mathbb{E}_{((x, c), y) \sim \mathcal{D}_{\text{clean}}} \left[ \mathcal{L}(f_{\theta}(x, c), y) \right] \nonumber \\
&+ \mathbb{E}_{((x', c), y^*) \sim \mathcal{D}_{\text{poison}}} \left[ \mathcal{L}(f_{\theta}(x', c), y^*) \right].
\end{align}

\section{Evaluation}
\subsection{Evaluation Settings}
\noindent \textbf{Dataset.}
Experiments are conducted on the KITTI dataset~\cite{geiger2013vision} and a newly-collected real-world physical dataset. We use the KITTI dataset, comprising 3,712 training and 3,769 validation samples, solely for training poisoned models. In contrast, physical-world threat validation is performed on newly collected real-world physical data, which contains approximately 500 synchronized Camera-LiDAR samples with triggers at varied poses and distances.

\noindent  \textbf{Physical Data Collection and Calibration.} 
To ensure our physical evaluation is both rigorous and comparable to existing benchmarks, our data collection process follows the sensor configuration guidelines of the KITTI dataset. We mounted a LiDAR sensor above a camera (as shown in Figure~\ref{fig:phys_setup}) and performed a two-stage calibration. First, intrinsic camera parameters (focal length, principal point, distortion) were estimated using MATLAB's Camera Calibration Toolbox~\cite{matlabcalib} with a standard checkerboard pattern. Second, the extrinsic transformation between the camera and LiDAR was determined by manually aligning point clouds to image features with the SensorsCalibration toolbox~\cite{pjlab_adg_sensorscalibration}.
Using this calibrated setup, we recorded clean data (without triggers) and poisoned data (with physical triggers attached to a vehicle). To thoroughly evaluate robustness, the data were captured under diverse environmental conditions over a wide range of distances and viewing angles. All collected samples were manually annotated in CVAT~\cite{cvat} and formatted to the KITTI standard.

\noindent \textbf{Baselines \& Attack Setup.}
We evaluate our attack, MOBA, against three state-of-the-art physical backdoor attacks: Zhang’s attack~\cite{zhang2022towards}, BadLiDet~\cite{li2023badlidet}, and BadFusion~\cite{chaturvedi2024badfusion}. In our experimental setup, we incorporate transformations, point sampling, and regularization to promote natural robustness during training, while maintaining a poisoning rate of 15\%. To ensure a fair comparison, we adapt each baseline to use our stealthy ``Baby on Board'' rectangular backdoor trigger. Each method is re-implemented in accordance with the procedures outlined in their respective papers. For BadFusion, to maintain consistency with MOBA and other baselines, we embed the digital LiDAR trigger into the point cloud alongside the corresponding camera images.
The attacks are tested on popular LiDAR-only models (SECOND~\cite{yan2018second}, PointPillars~\cite{lang2019pointpillars}, VoxelNet~\cite{zhou2018voxelnet}) and a Camera-LiDAR fusion model (MVX-Net~\cite{sindagi2019mvxnet}).

\noindent \textbf{Evaluation Metrics.}
We use three standard metrics:
(1) Clean mAP: The model's mean Average Precision on clean data, measuring its baseline performance.
(2) Attack Success Rate (ASR): The percentage of poisoned samples where the backdoor is successfully activated. For resizing attacks, ASR is defined as the proportion by which the bounding box sizes are reduced when the trigger is activated. In the case of disappearance attacks, ASR represents the proportion of bounding boxes that vanish upon trigger activation.
(3) Poisoned mAP: The mAP on poisoned data; a low value indicates the attack successfully degraded performance.

\begin{figure}[t]
    \centering
    \includegraphics[width=\columnwidth]{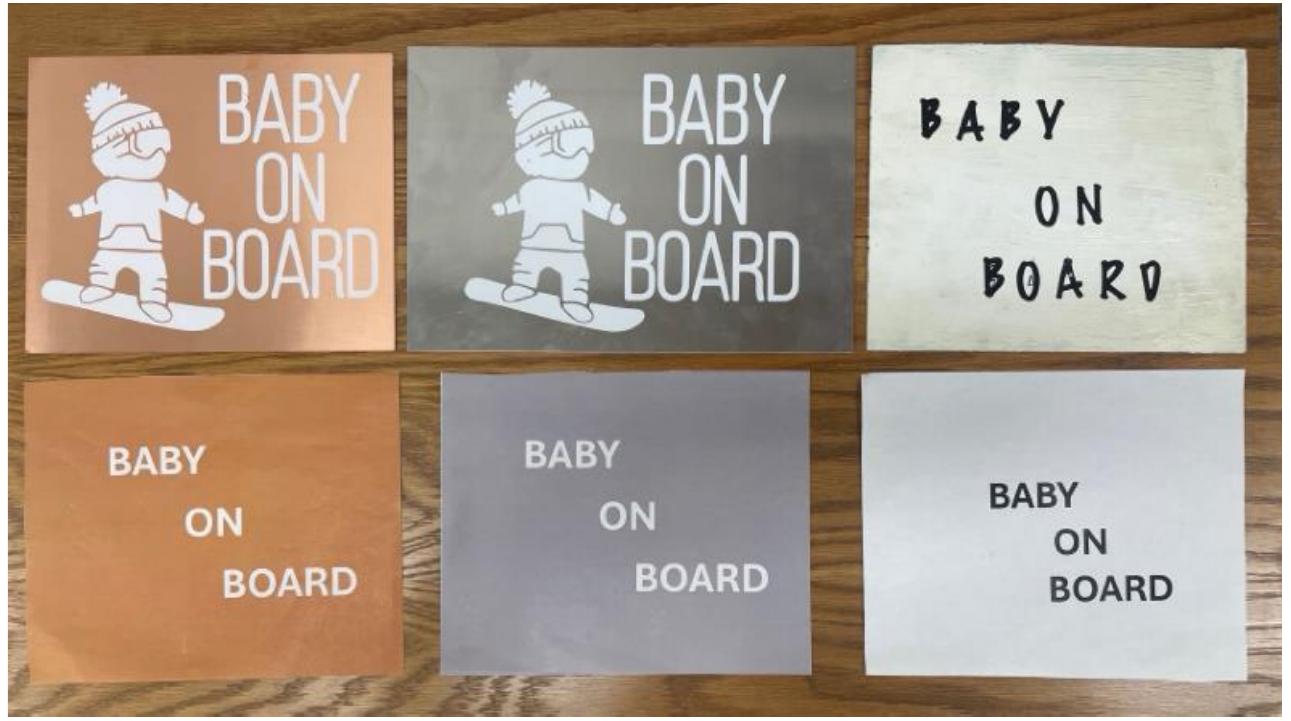}
    \caption{Examples of different ``Baby on Board'' physical triggers used in the evaluation. The top row shows triggers made of copper (left), aluminum (middle), and TiO$_2$ (right) materials. The bottom row includes similar ``Baby on Board'' triggers printed on paper materials.}
    \label{fig:physical_triggers}
\end{figure}

\begin{table}[t]
\caption{Performance comparison of the proposed MOBA against state-of-the-art attacks. We report the performance on our collected physical trigger dataset under the resizing bounding box attack across three LiDAR-based detection models. Here, AR represents the Angle Robustness component that computes intensity, and DR represents the Distance Robustness component.}
\small
\centering
\resizebox{\columnwidth}{!}{%
\begin{tabular}{llcc}
\toprule
\textbf{Models} & \textbf{Attack} & 
\textbf{Poison mAP (\%)$\downarrow$} & 
\textbf{ASR (\%)$\uparrow$} \\
\midrule
\multirow{6}{*}{VoxelNet} 
 & BadLiDet             & 71.56 & 49.72 \\
 & Zhang's attack           & 74.58 & 43.28 \\
 & MOBA W/o AR       & 19.83 & 84.35 \\
 & MOBA W/o DR       & 30.86 & 79.59 \\
 & \textbf{MOBA}     & 9.45 & 93.87 \\
\midrule
\multirow{6}{*}{SECOND} 
 & BadLiDet             & 78.11 & 40.17 \\
 & Zhang's attack          & 86.83 & 34.29 \\
 & MOBA W/o AR       & 10.55 & 57.14 \\
 & MOBA W/o DR       & 0.01 & 79.59 \\
 & \textbf{MOBA}     & 0.82 & 94.00 \\
\midrule
\multirow{6}{*}{PointPillars} 
 & BadLiDet             & 71.38 & 46.17 \\
 & Zhang's attack        & 75.75 & 39.51 \\
 & MOBA W/o AR       & 0.20 & 53.06 \\
 & MOBA W/o DR      & 12.01 & 85.68 \\
 & \textbf{MOBA}    & 7.07 & 90.23 \\
\bottomrule
\end{tabular}
}
\label{tab:main_baseline_results}
\end{table}

\subsection{Main Evaluation Results}
Our primary evaluation assesses the real-world effectiveness of MOBA against existing methods using our physically collected dataset. The results presented in Table~\ref{tab:main_baseline_results} demonstrate MOBA's superiority. While baseline methods like Zhang’s attack and BadLiDet struggle significantly in physical deployment, achieving low average ASRs of 49.72\% and 43.28\%, respectively, MOBA achieves an outstanding average ASR of 92.7\% across all LiDAR-only models, while maintaining a clean mAP of approximately 90\%. These results show that by neglecting physical-world material properties such as LiDAR intensity, conventional digital triggers fail to reliably transfer to the physical domain.

The importance of MOBA's physically grounded design is further validated by the results from our two ablated variants. Removing either the Intensity simulation using Angle Robustness component (AR) or the Distance Robustness component (DS) causes the ASR to drop by approximately 10–15\%. While these ablated versions still outperform the baselines, their reduced effectiveness highlights that each component of MOBA is critical to its success, working in concert to bridge the digital-to-physical gap.

Furthermore, MOBA's robustness extends to multimodal architectures. As shown in Table~\ref{tab:mvxnet_results}, MOBA achieves an exceptional 95.91\% ASR on the MVX-Net fusion model, decisively outperforming BadFusion’s 63.26\% ASR. These findings underscore the generalizability and robustness of MOBA across different model architectures and modalities.

\begin{table}[t]
\caption{Performance comparison of the proposed MOBA against state-of-the-art methods and baselines for the LiDAR-Camera fusion model MVX-Net on our collected physical trigger dataset under resizing bbox attack.}
\small
\centering
\begin{tabular}{lccc}
\toprule
\textbf{Method} & \textbf{Poison mAP (\%)$\downarrow$} & \textbf{ASR (\%)$\uparrow$} \\
\midrule
BadFusion           & 12.15 & 63.26 \\
MOBA W/o AR       & 2.80 & 81.63 \\
MOBA W/o DR      & 26.41 & 79.59 \\
\textbf{MOBA}    & 4.80 & \textbf{95.91} \\
\bottomrule
\end{tabular}
\label{tab:mvxnet_results}
\end{table}

\subsection{Ablation Study}
We analyze the effectiveness of MOBA under different material triggers, the impact of intensity estimation, the poisoning rate, and different attack objectives. Additional experiments on the impact of poisoning rate, distance and viewing angle robustness are provided in Appendix~\ref{appendix:experiments}

\noindent \textbf{Different Material Triggers:}
To evaluate the impact of material properties on trigger effectiveness, we compare MOBA’s performance using physically reflective materials (TiO$_2$ paint, copper, and aluminum) against visually similar but non-reflective paper-based alternatives (WhiteBas, CuBas, AluBas). As shown in Table~\ref{tab:material_trigger_results}, the TiO$_2$ trigger achieves the highest ASR (95.91\%), followed by copper (78.43\%) and aluminum (37.25\%). In contrast, paper-based versions yield significantly lower ASRs as compared to TiO$_2$ paint (around 50\%), confirming that visual similarity alone is insufficient. These results underscore the importance of using reflective materials like TiO$_2$ to ensure strong LiDAR returns and reliable trigger activation in real-world settings.

\begin{table}[t]
\caption{ASR comparison for different physical trigger materials and their paper-based counterparts.}
\small
\centering
\begin{tabular}{lc}
\toprule
\textbf{Trigger Material} & \textbf{ASR (\%) $\uparrow$} \\
\midrule
\textbf{Titanium Dioxide (TiO$_2$)}         & \textbf{95.91} \\
Copper (Cu)            & 78.43 \\
Aluminum (Alu)       & 37.25 \\
Paper (Cu)   & 50.98 \\
Paper (White)      & 46.05 \\
Paper (Alu)        & 49.01 \\
\bottomrule
\end{tabular}
\label{tab:material_trigger_results}
\end{table}
\noindent \textbf{Impact of Intensity Estimation:}
To quantify the impact of different LiDAR intensity modeling strategies on MOBA’s ASR, we compare four configurations: (1) no intensity, (2) random intensity values, (3) fixed intensity of 0.5, and (4) BRDF-based intensity (see Section~\ref{sec:proposed_meth}). Table~\ref{tab:intensity_ablation_results} shows that BRDF-based intensity achieves the highest ASR (95.91\%) by best replicating real-world reflectance. Removing or inaccurately modeling intensity significantly reduces ASR, underscoring the importance of realistic LiDAR simulation.
\begin{table}[t]
\caption{ASR of digital triggers with different LiDAR intensity simulation settings on MVX-Net.}
\small
\centering
\begin{tabular}{lc}
\toprule
\textbf{Intensity Setting} & \textbf{ASR (\%) $\uparrow$} \\
\midrule
\textbf{BRDF-based intensity}      & \textbf{95.91} \\
Fixed (0.5)          & 79.59 \\
Random               & 91.83 \\
No Intensity         & 81.63 \\
\bottomrule
\end{tabular}
\label{tab:intensity_ablation_results}
\end{table}

\noindent \textbf{Impact of Different Attack Objectives:}
We evaluate MOBA’s flexibility across different attack goals: object resizing and disappearance. Table~\ref{tab:attack_goal_voxelnet} shows that MOBA achieves high ASRs for both objectives on VoxelNet. Resizing yields the highest ASR at 93.87\%, while disappearance is more challenging but still achieves strong attack success and reduces mAP on poisoned samples by over 95\%. These results confirm MOBA’s adaptability to a range of threat scenarios.  
For results on MVX-Net, please refer to Table~\ref{tab:attack_goal_mvxnet} in the Appendix.

\begin{table}[H]
\caption{Impact of different attack objectives on MOBA for VoxelNet.}
\small
\centering
\begin{tabular}{lc}
\toprule
\textbf{Attack Objective} & \textbf{ASR (\%) $\uparrow$} \\
\midrule
Resizing       & 93.87 \\
Disappearance  & 95.33 \\
\bottomrule
\end{tabular}
\label{tab:attack_goal_voxelnet}
\end{table}

\section{Conclusion}
This paper presents \textbf{MOBA}, a novel Material-Oriented Backdoor Attack targeting 3D object detection systems in autonomous driving. MOBA is the first physical backdoor framework to explicitly leverage the material-dependent reflectivity characteristics of LiDAR sensors, enabling the design of robust and stealthy physical triggers. By simulating LiDAR-consistent intensity values in the digital domain and aligning the trigger's spatial characteristics with real-world constraints, MOBA achieves high attack success rates across both LiDAR-only and camera-LiDAR fusion models.
Comprehensive experiments validate the effectiveness and practicality of MOBA under diverse attack objectives, material configurations, and poisoning rates. In particular, the use of TiO$_2$-based triggers and precise intensity estimation significantly enhances the transferability of digital attacks to the physical world. Moreover, MOBA remains highly effective even at low poisoning rates, highlighting its viability in practical deployment scenarios.
Overall, this work reveals a previously unexplored vulnerability in LiDAR-based 3D object detection pipelines and highlights the critical need to consider defenses that account for material-level properties in real-world environments.

\section{Acknowledgment}
This work was supported in part by the National Science Foundation under NSF Award 2444389.

\bibliography{aaai2026}

\clearpage
\appendix
\appendix
\section{Appendix}

This appendix provides supplementary details on material modeling, LiDAR intensity simulation, and additional experimental results discussed in the main paper.




\subsection{Material Properties and Selection Analysis}
\label{appendix:selection}

To apply our material selection objective (Eq.~\ref{eq:materialselection} in the main text), we first gathered the necessary optical properties for each candidate material at the standard automotive LiDAR wavelength of $905$ nm. These properties, sourced from established optical databases and material science literature, are summarized in Table~\ref{tab:material_properties}.

\begin{table*}[t]
\centering
\small
\caption{Optical Properties of Candidate Materials at 905 nm. The complex refractive index is given in the format $n + ik$, where $n$ is the refractive index and $k$ is the extinction coefficient.}
\label{tab:material_properties}
\begin{tabular}{lcccc}
\toprule
\textbf{Material} & \textbf{Refractive Index ($n_M$)} & \textbf{Diffuse Reflectance ($\rho$)} & \textbf{Surface Roughness ($\sigma$)} \\
\midrule
Aluminum  & $1.43 + 8.33i$ & $\sim 0.10$ & $\sim 0.05$  \\
Copper & $0.23 + 6.09i$ & $\sim 0.08$ & $\sim 0.05$  \\
Paper & $\sim 1.50$ & $\sim 0.75$ & $\sim 0.80$ \\
\textbf{Titanium Dioxide (TiO$_2$)} & \textbf{2.51} & \textbf{$\sim 0.95$} & \textbf{$\sim 0.70$} \\
\bottomrule
\end{tabular}
\end{table*}

Using these properties, we computed the material selection score ($M^*$) for each candidate across a range of incident angles ($\theta_i$) from 0$^\circ$ (normal incidence) to 80$^\circ$ (grazing angle), with a weighting factor of $\alpha=0.2$ to prioritize diffuse reflectance. The averaged scores are presented in Table~\ref{tab:material_scores}.

\begin{table*}[t]
\centering
\small
\caption{Material Selection Score ($M^*$) Evaluation}
\label{tab:material_scores}
\begin{tabular}{lcccc}
\toprule
\textbf{Material} & \textbf{Avg. $R_{\text{specular}}$} & \textbf{Avg. $R_{\text{diffuse}}$} & \textbf{Avg. $M^*$ Score} & \textbf{Rank} \\
\midrule
Aluminum & 0.92 & 0.03 & 0.21 & 2 \\
Copper & 0.94 & 0.02 & 0.20 & 3 \\
Paper & 0.04 & 0.20 & 0.17 & 4 \\
\textbf{Titanium Dioxide (TiO$_2$)} & \textbf{0.18} & \textbf{0.28} & \textbf{0.26} & \textbf{1} \\
\bottomrule
\end{tabular}
\end{table*}

The results clearly show that \textbf{Titanium Dioxide (TiO$_2$)} achieves the highest score. While polished metals have high specular reflectance, their performance degrades sharply at non-normal angles, resulting in a lower overall score. TiO$_2$'s combination of high diffuse reflectance and optimal surface roughness makes it the most robust choice for generating a consistent LiDAR signature across diverse real-world conditions.

\subsection{Derivation of Environmental Robustness}
\label{appendix:environmental}

\paragraph{Water Accumulation and Hydrophobicity}

Water accumulation alters the surface refractive index as:
\begin{equation}
n_{\text{eff}} = f \cdot n_{\text{water}} + (1 - f) \cdot n_2,
\end{equation}
where \( f \) is the fractional water coverage and \( n_{\text{water}} = 1.33 \). For hydrophobic materials like TiO$_2$, \( f \approx 0 \), ensuring:
\begin{equation}
R_{\text{wet}} \approx R_{\text{dry}}.
\end{equation}

\paragraph{LiDAR Beam Divergence and Planar Geometry}

LiDAR beam divergence is defined as:
\begin{equation}
\theta_B = \frac{\lambda}{D},
\end{equation}
with divergence augmented by curved surfaces:
\begin{equation}
\theta' = \theta_B + \Delta \theta,
\end{equation}
For planar surfaces:
\begin{equation}
\theta'_{\text{planar}} \approx \theta_B, \quad \text{and} \quad \rho_{\text{planar}}(d) \gg \rho_{\text{curved}}(d),
\end{equation}
ensuring better point cloud density over varying distances.

\subsection{Experimental Results}
\label{appendix:experiments}


\noindent \textbf{Robustness to Distance.}
To evaluate the effectiveness of MOBA’s distance robustness design, we test its performance at varying distances from the target vehicle: close, medium, and far. As shown in Table~\ref{tab:distance_robustness_results}, the full MOBA framework maintains a high and stable ASR across all distances. In contrast, removing the distance robustness component leads to a significant performance drop, especially at medium distances, where ASR falls to just 30\%. This demonstrates that without adaptive scaling, the trigger's point cloud representation becomes inconsistent across depths, breaking the learned backdoor association. These results confirm that distance robustness is essential for ensuring attack reliability in dynamic, real-world conditions.

\begin{table}[H]
\caption{ASR (\%) comparison from different viewing angles to evaluate the angle-independent BRDF approximation.}
\small
\centering
\begin{tabular}{lc}
\toprule
\textbf{Viewing Angle} & \textbf{ASR (\%)} \\
\midrule
Left Angle  & 91.67 \\
Straight    & 88.88 \\
Right Angle & 96.42 \\
\bottomrule
\end{tabular}
\label{tab:angle_robustness_results}
\end{table}

\noindent \textbf{Robustness to Viewing Angle.}
To assess the real-world impact of our angle robustness design via BRDF approximation, we evaluate attack success from three different viewing perspectives relative to the trigger: left, center, and right. As shown in Table~\ref{tab:angle_robustness_results}, MOBA achieves consistently high ASR across all angles, averaging 92.32\%. This stability validates our strategy of abstracting away the precise viewing angle during digital simulation, demonstrating that the learned trigger is resilient to natural variations in perspective encountered during real-world driving.

\begin{table}[H]
\caption{ASR (\%) comparison at different distances to evaluate the distance-robustness component of MOBA.}
\small
\centering
\begin{tabular}{lccc}
\toprule
\textbf{Method} & \textbf{Close} & \textbf{Medium} & \textbf{Far} \\
\midrule
MOBA (Full)           & 100.00 & 80.00 & 100.00 \\
MOBA W/o AR           & 89.00  & 50.00 & 89.00  \\
MOBA W/o DS         & 69.00  & 30.00 & 69.00  \\
\bottomrule
\end{tabular}
\label{tab:distance_robustness_results}
\end{table}

\noindent \textbf{Impact of Poisoning Rate on VoxelNet:}
We analyze the trade-off between poisoning rate and ASR to assess MOBA’s practicality under limited data poisoning. Table~\ref{tab:poisoning_rate_voxelnet} reports ASRs for poisoning rates of 5\%, 10\%, and 15\% on VoxelNet. Even with only 5\% poisoning, MOBA achieves a 70\% ASR, which increases to 93.87\% with 15\% poisoning. These results demonstrate that MOBA remains effective even under limited poisoning—an essential consideration for realistic threat scenarios where large-scale poisoning may be infeasible.  
For results on MVX-Net, please refer to Table~\ref{tab:poisoning_rate_mvxnet} in the Appendix.
\begin{table}[H]
\caption{Impact of poisoning rates on ASR for VoxelNet.}
\small
\centering
\begin{tabular}{lc}
\toprule
\textbf{Poisoning Rate} & \textbf{ASR (\%) $\uparrow$} \\
\midrule
5\%     & 70.00     \\
10\%    & 90.00    \\
15\%    & 93.87  \\
\bottomrule
\end{tabular}
\label{tab:poisoning_rate_voxelnet}
\end{table}

\noindent \textbf{Impact of Poisoning Rate on MVX-Net:}
We investigate the effect of varying the poisoning rate on the attack success rate (ASR) for MVX-Net to evaluate MOBA's efficiency under constrained poisoning conditions. Table~\ref{tab:poisoning_rate_mvxnet} in the Appendix shows that even at a low poisoning rate of 5\%, MOBA achieves an ASR of 65\%. Increasing the poisoning rate to 10\% and 15\% leads to further improvements, reaching an ASR of 95.91\% at 15\%. These results confirm that MOBA is effective and practical even when only a small fraction of the training data is poisoned, making it a viable threat model for real-world scenarios where large-scale poisoning may be difficult.

\begin{table}[H]
\caption{Impact of different attack objectives on MOBA for MVX-Net.}
\centering
\begin{tabular}{lc}
\toprule
\textbf{Attack Objective} & \textbf{ASR (\%)} \\
\midrule
Resizing       & 95.91 \\
Disappearance  & 58.33 (99.77) \\
\bottomrule
\end{tabular}
\label{tab:attack_goal_mvxnet}
\end{table}

\noindent \textbf{Impact of Attack Objectives on MVX-Net:}
To demonstrate the versatility of MOBA, we test it under two distinct attack objectives: object resizing and disappearance. As reported in Table~\ref{tab:attack_goal_mvxnet} (Appendix), the resizing objective yields the highest ASR of 95.91\%. Although disappearance is inherently more challenging, the attack still succeeds with a 58.33\% ASR, while reducing the mean average precision (mAP) on poisoned samples by 99.77\%. These findings validate that MOBA can be adapted to serve different adversarial goals depending on the attacker's intent, including subtle manipulations like reducing object size or more aggressive objectives like complete disappearance.

\begin{table}[H]
\caption{Impact of poisoning rate on ASR for MVX-Net.}
\centering
\begin{tabular}{lc}
\toprule
\textbf{Poisoning Rate} & \textbf{ASR (\%)} \\
\midrule
5\%     & 65     \\
10\%    & 69     \\
15\%    & 95.91  \\
\bottomrule
\end{tabular}
\label{tab:poisoning_rate_mvxnet}
\end{table}

\end{document}